# Self-Adapting Recurrent Models for Object Pushing from Learning in Simulation

Lin Cong*, Michael Görner, Philipp Ruppel, Hongzhuo Liang,
Norman Hendrich, Jianwei Zhang

*Abstract*— Planar pushing remains a challenging research topic, where building the dynamic model of the interaction is the core issue. Even an accurate analytical dynamic model is inherently unstable because physics parameters such as inertia and friction can only be approximated. Data-driven models usually rely on large amounts of training data, but data collection is time consuming when working with real robots.

In this paper, we collect all training data in a physics simulator and build an LSTM-based model to fit the pushing dynamics. Domain Randomization is applied to capture the pushing trajectories of a generalized class of objects. When executed on the real robot, the trained recursive model adapts to the tracked object's real dynamics within a few steps. We propose the algorithm Recurrent Model Predictive Path Integral (RMPPI) as a variation of the original MPPI approach, employing state-dependent recurrent models. As a comparison, we also train a Deep Deterministic Policy Gradient (DDPG) network as a model-free baseline, which is also used as the action generator in the data collection phase. During policy training, Hindsight Experience Replay is used to improve exploration efficiency. Pushing experiments on our UR5 platform demonstrate the model's adaptability and the effectiveness of the proposed framework.

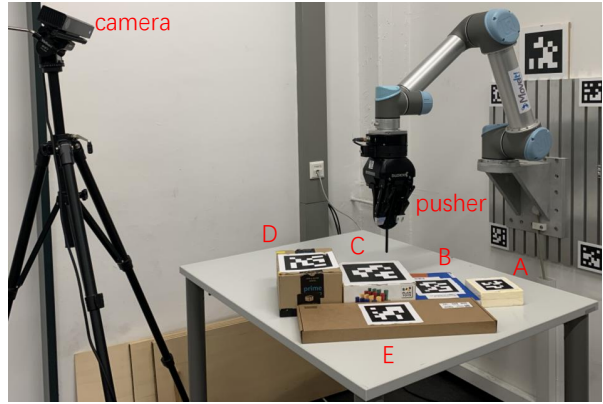

Fig. 1. Our experiment platform consists of an UR5 robot with Robotiq 3-finger hand which grasps the 3D-printed vertical pusher rod. Its cylindrical part, designed to touch and push the moving object, is 6 mm in diameter. The five objects marked from A to E are test objects with different physical properties (sliding friction, rotating friction, mass, and size). Object position and motion are tracked using AprilTag markers and the camera.

## I. INTRODUCTION

Planar object pushing with a single contact is a typical underactuated instance of robot manipulation. The uncertainty of different physics parameters and the pressure distribution makes it difficult to build a precise motion model for real interactions. However, learning a data-driven model [1], [2] is an effective method due to the stochastic nature [3] of the pushing process. Besides the model-based method, end-to-end policy learning by mapping from state space to action space directly is also becoming a trend [4], [5]. A primary defect of such model-free methods remains the difficulty to collect sufficient manipulation experience on real robots. Notorious instability of robot behavior during the exploration phase limits the feasibility of training. To avoid physical training, policies can be trained in simulation environments. In order to deploy such policies successfully in the real world though, the training framework has to overcome the well-known gap between simulation and reality. In [4], Andrychowicz et al., use domain randomization to transfer their simulation result to a real environment successfully.

In this paper, we use an LSTM-based model to predict the motion of objects with unknown parameters and apply Model Predictive Control (MPC) [6] as control strategy to push the objects to target poses with a single point of contact. Recurrent Neural Networks (RNNs) have long been used for fitting nonlinear dynamic models. With the distinctive ability to recognize patterns in sequences of data, the LSTM module is chosen to fit the object motion dynamics during the pushing process.

MPPI [7], [8], as a typical Model Predictive Control method, is investigated for optimizing nonlinear system model control. We focus on the planar pushing problem and set the task as pushing different objects with unknown physics parameters to different target poses. This paper investigates the use of an LSTM-based dynamic model, trained fully in simulation, to predict motion in a real environment. We use Domain Randomization in simulation to change the physics parameters of the target objects, including mass, size, sliding friction, and rotational friction. In the real world pushing process, the trained LSTM continuously observes the actual object motion and self-adapts to the object physics within a few (less than 5) pushing steps.

With this paper we contribute multiple insights:

- We propose *RMPPI*, integrating traditional MPPI with recurrent state-dependent models.
- We demonstrate that a state-dependent model trained from simulation can be effective for predicting the motion of real objects without further fine-tuning.
- We show that a model-free policy trained with a pro-

*Corresponding author ⟨cong@informatik.uni-hamburg.de⟩
Authors are with the Department of Informatics, University of Hamburg.
*This research was partially funded by the German Research Foundation (DFG) and the National Science Foundation of China in project Crossmodal Learning, TRR-169 and the CSC (China Scholarship Council)

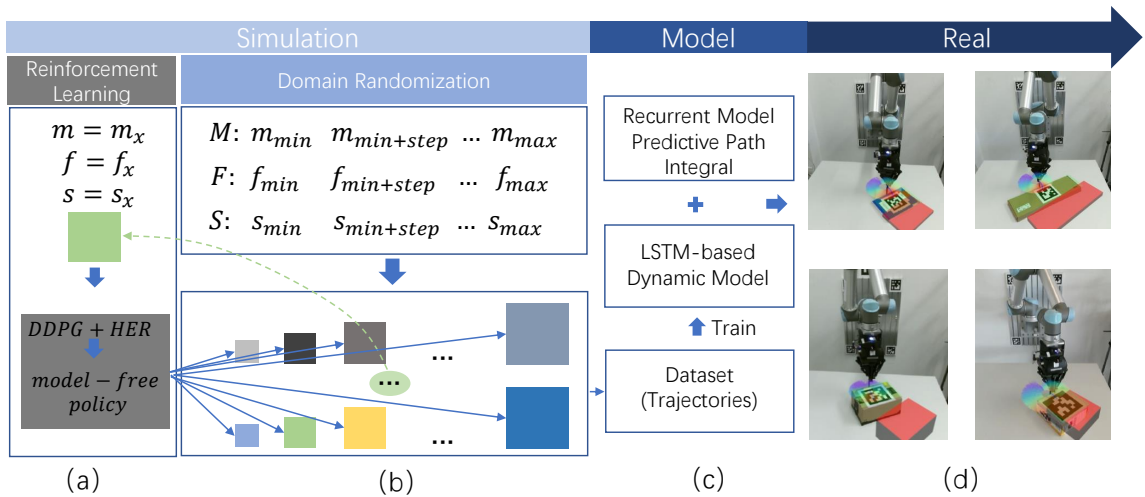

Fig. 2. Overview: (a) A model-free generator policy (baseline also) is trained using DDPG and HER to push a prototypical object in a physics simulation. (b) The generator policy is used to explore randomized object with biased object-interactions in the simulator. $M$, $F$ and $S$ represent mass, friction and size respectively. (c) All sampled trajectories are used to train an LSTM-based dynamic model. (c-d) The model provides multi-step rollouts for a Recurrent Model Predictive Path Integral controller to actuate the robotic system.

totypical object can efficiently bias the exploration of novel objects for subsequent model learning.

## II. RELATED WORK

### A. Robot Manipulation

Planar pushing is a popular research topic. Pushing can be taken as a pre-action before some purposive action, such as grasping. In general, the solutions to most kinds of robot manipulation can be divided into model-based and model-free. The model-based method, combined with Model Predictive Control theory, is widely used in various robotic tasks. An analytical method is always the right choice when the parameters of the specific object are known for sure. However, getting a stable dynamic model in a real environment is never an easy task.

After Mason introduced the modeling problem of pushing decades ago [9], a lot of research has been done around the modeling of planar pushing. In recent years, data-driven methods for building a dynamic model for complicated nonlinear systems have increasingly attracted attention from researchers. Gaussian Process (GP) regression [10] and deep neural network [11] are two typical data-driven modeling methods widely used.

Given a correct general model, the feedback mechanism of control theory can achieve satisfactory control results. In many robot tasks that involve complex interactions like manipulating a cube with a robotic hand [4] or the problem of using tools [12], it is impossible to model precisely for every contact situation. In these cases, a model-free method such as deep reinforcement learning, which can map from high-dimensional state space to action space directly, is more suitable. However, deep reinforcement learning suffers from the necessity for large amounts of experience data, which is difficult to obtain for real robots. Recently some literature has also been combining model-based with model-free methods [13] to improve exploring efficiency.

### B. Pushing Model

To predict an object's motion, a traditional method is to describe the dynamics as an analytical model [14]. Parameters in the analytical model have specific physical meanings, therefore, the model can be easily transferred to similar problems given specific system parameters. However, generalizing the analytical model to unknown objects is difficult, especially when the properties of the object, such as friction and inertia, are hard to get. Recently, data-driven models [15] have successfully been used for predicting object motion and outperform analytical models. In [16], Bauza *et al.* use GP to fit the dynamic model with less than 100 data points from the real environment and get a very stable trajectory tracing performance. The approach is very efficient. However, the model cannot deal with new objects. In [17], Arruda *et al.* use Gaussian Process Regression and an Ensemble of Mixture Density Networks and get both mean and variance, which is a good evaluation of uncertainty in their predictions.

### C. Simulation to Reality

Training robots in a simulated environment is convenient as it is easy to collect large amounts of data, and we do not have to consider safety problems during the training process. However, transferring results or policies trained in simulation to a real environment is always challenging. Domain randomization [18] is a useful technique for bridging the gap between simulation and real environment. In [19], Peng *et al.* use Domain Randomization to bridge the gap by randomizing the dynamics parameters in simulation during training. The result shows excellent robustness. However, their algorithm relies on a high-performance machine, which takes approximately 8 hours for the policy to do exploration on a 100 core cluster. In this paper, it takes only 2 hours to generate adequate training data on an 8 core laptop.

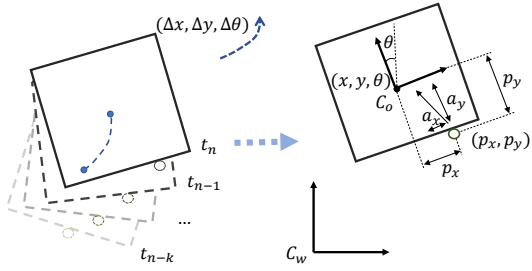

Fig. 3. The figure illustrates a planar pushing system by showing the time sequence and variable representations in the pushing problem description. $C_w$ and $C_o$ denote word coordinate and object coordinate, respectively.

## III. METHOD

### A. Self-Adapting Recurrent Model

Humans can approximate dynamic models from several pushing steps during the interaction with an object and choose the suitable pushing direction and velocity given a target pose. Inspired by this memory learning process, we choose an LSTM module as the main part of our dynamic model. Figure 3 illustrates the planar pushing system, where $p_x, p_y$ and $a_x, a_y$ denote the pusher's position and the action in the object's frame. These variables are taken as inputs for the model in [1], [16]. $x, y, \theta$ denote the position of the object's geometric center and the object's rotation in the world frame. $\Delta x, \Delta y, \Delta \theta$ are the corresponding increments relative to a previous moment in the object's frame. In theory, the dynamic property of the object can be inferred from its historical motion trajectory. In this paper, the state increments $\Delta x, \Delta y, \Delta \theta$ instead of absolute trajectory data $x, y, \theta$ are adopted as part of the input to eliminate the influence of possible data distribution drift caused by different world coordinate origins:

$$\Delta x_t, \Delta y_t, \Delta \theta_t = \begin{cases} 0, & \text{if } t = 0 \\ [x_t - x_{t-1}, y_t - y_{t-1}, \theta_t - \theta_{t-1}], & \text{if } t > 0 \end{cases} \quad (1)$$

In summary, at moment $t$, a tensor representing the current state of the system can be denoted as $\mathcal{S}_t$:

$$\mathcal{S}_t = [\Delta x_t, \Delta y_t, \Delta \theta_t, p_{x_t}, p_{y_t}, a_{x_t}, a_{y_t}] \quad (2)$$

We assume that the model $\mathcal{F}$ relies on the previous system state of $k+1$ time steps as the input to predict the motion in the object's frame for the next time step:

$$\Delta x_{t+1}, \Delta y_{t+1}, \Delta \theta_{t+1} = \mathcal{F}([\mathcal{S}_{t-k}, \mathcal{S}_{t-k+1}, ..., \mathcal{S}_t]) \quad (3)$$

After all necessary states are preprocessed and stacked as input, it is fed into the recurrent module (2-layer LSTM with 128 and 64 units). By preserving sequential information in the recurrent network's hidden state, LSTM module achieves self-adapting to the real dynamics and output a tensor of 64 recurrent features. Then the motion predictor (2-layer fully connected network) takes the recurrent tensor as input and yields the motion prediction in the object's frame. Figure 4 illustrates the details of the recurrent model. The dropout is set to 0.5 for both layers of the LSTM. We train the model through stochastic gradient descent on the $L_2$ loss between the prediction from the network and the real object motion using the Adam [20] optimizer.

### B. Recurrent Model Predictive Path Integral

MPPI [7], [8] has already been successfully applied in [21] for complicated robot manipulation problems in simulation. In order to endow the original MPPI with a memory mechanism, we add a history buffer $\mathcal{H}$ into the algorithm. Algorithm 1 gives details of the whole framework. The results from our real experiments prove the effectiveness of RMPPI.

---

**Algorithm 1:** Recurrent Model Predictive Path Integral (RMPPI)

---

**Given:**

$N$: Number of sampled action sequences;
$T$: Number of time steps to roll out;
$K + 1$: Number of initial action steps;
$F$ : $HistoryBuffer \times Action \rightarrow State$: Dynamic Model;
$C$ : $State \times Environment \rightarrow Cost$: Cost function;
$(u_0, u_1, ..., u_K)$: Initial control sequence;

***LSTM Warm Up Stage:***
$\mathcal{H} =$empty $HistoryBuffer$;
**for** $t \leftarrow 0$ **to** $K$ **do**
  $\mathcal{S}_t \leftarrow GetState(t)$;
  append $\mathcal{S}_t$ to $\mathcal{H}$;
  $RobotAction(u_t)$;

***Model Based Action Stage:***
**while not** $TaskCompleted()$ **do**
  Sample rollout actions $\mathcal{U}_{1...T}^{1...N}$;
  $\mathcal{H}^{1...n} = \mathcal{H}$;
  **for** $t \leftarrow 0$ **to** $T - 1$ **do in parallel**
    $\mathcal{S}_{t+1}^n = F(\mathcal{H}^n, \mathcal{U}_t^n)$;
    append $\mathcal{S}_{t+1}^n$ to $\mathcal{H}^n$;
    $\mathcal{C}_{t+1}^n = C(\mathcal{S}_{t+1}^n, Env)$;
  $\mathcal{U}_{1,...,n}^* = MPPI(\mathcal{C}, \mathcal{U})$;
  $RobotAction(\mathcal{U}_1^*)$;
  $\mathcal{S}_t \leftarrow GetState(t)$;
  append $\mathcal{S}_t$ to $\mathcal{H}$;

---

Before arriving at reasonable predictions, the LSTM module needs several steps to warm up. At the start of each pushing episode, we give $K+1$ steps of the initial control sequence $(u_0, u_1, ..., u_K)$ to the robot, which is the '***LSTM Warm Up Stage***' in Algorithm 1. The length of the initial control sequence depends on the input sequence we need to feed into the network.

After this stage, the memory buffer stores adequate states for the model to predict the object's motion. Then the algorithm goes into the '***Model Based Pushing Stage***'. For each pushing step, $N$ pushing action sequences are sampled, and the LSTM prediction is calculated in parallel for all $N$ sample trajectories and recurrently for $T$ time steps to do the rollouts. The costs of each sampled action sequence are set as the distance between the current state and the target.

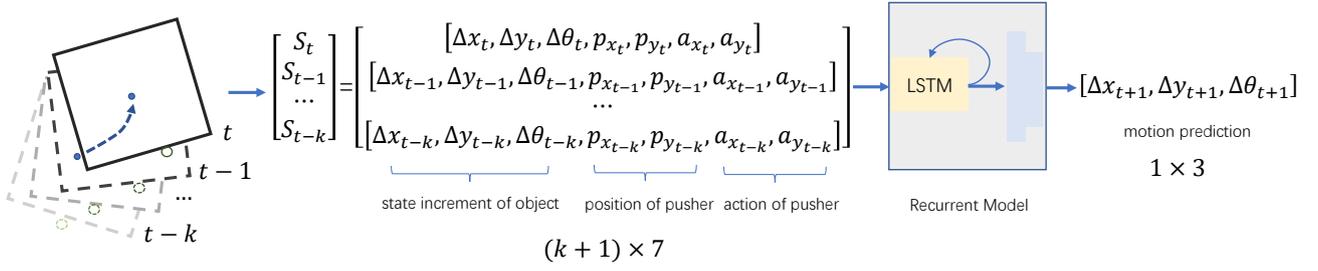

Fig. 4. The model architecture used in our experiments. The trajectory steps are preprocessed and stacked into the input sequence for the LSTM module, followed by a two-layer fully connected network (the blue bar in the figure). The output from the module is also a motion sequence, the last element of which is the motion prediction for the next time step. In figure we present only the last element instead of the whole output sequence.

$$\mathcal{C}_t = \|\mathcal{S}_t - \mathcal{T}_t\| \tag{4}$$

Then, using the path integral formula from MPPI, an optimal action list $(u^*_{t+1}, u^*_{t+2}, ..., u^*_{t+T})$ is computed. Only the first action $u^*_{t+1}$ is sent to the robot actuator and this process is repeated for every pushing step.

### C. Model-Free Pushing Baseline

In this paper, we also train a policy network that maps the system's current state to robot action directly using the classic off-policy Reinforcement Learning method DDPG [22]. The technique HER (Hindsight Experience Replay) [23], which can be combined with an arbitrary off-policy RL to improve the exploration efficiency, is applied during training. The initial idea of training this policy network is to do a comparative experiment as the model-free baseline. Both the results from simulation and real experiments show that the model-free method is valid only for the prototypical object used during the training process, but cannot be generalized to objects with different physical parameters. Even though the policy is not sufficient for achieving pushing tasks of a different object, during the study, we find that the policy can be used to collect pushing data for different objects. We will give details on the data collection process in the next section.

In our task, the pusher mounted on the robot arm is taken as the agent interacting with the stochastic environment $E$, which is the pushing task scenario. The agent's behavior is determined by policy $\pi$, which maps states to actions $\pi: \mathcal{S} \to \mathcal{A}$. In the task, we take sparse binary rewards and follow a Multi-Goal Reinforcement Learning framework in which the agent is told what to do using additional input [24]. We model it as a Markov decision process with state space:

$$\mathcal{S} = [X_o, Y_o, \theta_o,\ X_r, Y_r,\ X_g, Y_g, \theta_g] \tag{5}$$

in which the variables denote the current pose of the object, the robot end effector and the goal pose respectively. The action space is:

$$\mathcal{A} = [X_a, Y_a] \tag{6}$$

in which $X_a$ and $Y_a$ denote the action of the robot end effector. Other elements in the Markov decision process are: initial state distribution $p(s_0)$; transition dynamics $p(s_{t+1}|s_t, a_t)$, and sparse reward function:

$$r(s_t, a_t) = \begin{cases} -1, & \text{if goal is not achieved} \\ 1, & \text{if goal achieved} \end{cases} \tag{7}$$

The return of a state-action pair is defined as the sum of discounted future reward $R_t = \sum_{i=t}^{T} \gamma^{(i-t)} r(s_i, a_i)$ in which $\gamma$ is a discounting factor $\gamma \in [0, 1]$. The Bellman equation is used to describe the recursive relationship:

$$Q^\pi(s_t, a_t) = \mathbb{E}_{r_t, s_{t+1} \sim E}[r(s_t, a_t) + \gamma \mathbb{E}_{a_{t+1} \sim \pi}[Q^\pi(s_{t+1}, a_{t+1})]] \tag{8}$$

The greedy policy $\mu(s) = \arg\max_a Q(s, a)$ is applied in the framework.

### D. Domain Randomization Data Collection

Performing robotic learning in simulation has recently become especially promising for reaching human-level performance in various tasks such as using tools [12], object localization [18] and games like Atari [20]. The main advantage of learning in simulation is a faster, lower-cost data collection process. In this paper, we use Domain Randomization to randomize 300 objects with different physical parameters. With enough variability, an object from the real world may appear to be a variation from the randomized domains [19]. As represented in figure 2, objects are generated according to physics parameters from different domains in Gym [25], [26], after which the data collection process is performed on these objects. All the objects in simulation have the same size in height (2.5cm). Table I details the range of objects' parameters.

Instead of applying random actions to exploring object properties as in [2], [3], we use the trained model-free policy (baseline approach from the previous part) to generate actions to push all the randomized objects to goal poses. The DDPG policy is trained with a prototypical object with fixed parameters and cannot generalize to different objects. The specific parameter of the prototypical object is shown in table II as object P. As a result, this policy cannot push the random object to the target pose ideally and it will generate noisy actions, but the general movement tendency is towards the target pose. This kind of robot motion is especially suitable for exploring the objects' properties. During the whole exploration (data collection) process, the initial and

TABLE I
Dynamic Parameters and Their Ranges in Simulation

| Parameters | Range |
|---|---|
| Size | [8, 25] cm in length = width |
| Mass | [0.01, 0.8] kg |
| Sliding Friction Coefficient | [0.1, 1] |
| Rotation Friction Coefficient | [0.001, 0.01] |
| Damping Coefficient | [0.01, 0.015] |

TABLE II
Parameters of Experimental Objects

|   | Mass (kg) | Length (m) | Width (m) | Sliding Friction (N) |
|---|---|---|---|---|
| A | 0.016 | 0.116 | 0.116 | < 0.1 |
| B | 0.615 | 0.168 | 0.237 | ≈ 1.4 |
| C | 0.565 | 0.198 | 0.198 | ≈ 1.1 |
| D | 0.587 | 0.166 | 0.228 | ≈ 1.8 |
| E | 0.506 | 0.153 | 0.462 | ≈ 0.9 |
| P | 0.015 | 0.120 | 0.120 | ≈ 0.05 |

Object from A to E are objects in real experiments, P is the prototypical object used in the model-free policy training.

target poses of the objects are set randomly for going through as many different states of the object as possible. An episode is regarded as finished either when the object reaches the target pose or the 60th pushing step is executed by the robot. This data collection approach proves to be sufficient to explore objects with different parameters.

## IV. Experiments

### A. Robot Setup

We test our methods in the real world on a UR5 robot shown in figure 1. The robot is mounted to a wall above a table and holds a pushing rod (pusher) to interact with the objects. Object positions are measured using a camera and AprilTag markers [27]. Pushing commands are interpreted as Cartesian motion goals. These motion goals are translated by an online controller [28] into joint-space robot commands under kinematic and dynamic constraints. End-effector positions are restricted to a rectangular workspace above the table and the axis along the pushing rod is fixed in an upright position. Motions are constrained by velocity and acceleration limits.

### B. Benchmark Description

We propose a model-based method for pushing different objects, and the model is trained with all the data from the simulation. A model-free method, which is a deep policy network trained by DDPG combined with HER, is taken as the baseline. Both approaches are evaluated in real experiments. Through experiments, we find that the hyperparameters in RMPPI have significant impact on the pushing performance. The data analysis is detailed in the next section. Figure 5 shows selected steps from one pushing experiment using the square box (object C) in rviz. The pictures from each column represent robot and object states of the same time step. Green and red rectangles represent the current pose and target pose respectively. In each figure, the 1024 rollout action sequences of the pusher are shown. The color of each line is the normalized estimated cost of the action sequence, computed by the cost function we defined acting on the prediction results from the dynamic model. Red means a low-cost pushing direction, while blue stands for a high cost.

To demonstrate the adaptability of our model by predicting objects with unknown motion properties, we plot the trajectories of both object and pusher in Figure 6. (a) and (c) show the trajectories of object and pusher from pushing experiments with object E, (b) and (d) are from object C. Light gray squares represent the ending pose of each pushing episode. The three target poses are colored by red, blue, and green, respectively.

To compare the motion properties of different objects, during the experiment, we set the same target poses for different objects. From the trajectories of the pusher, we notice that the initial several pushing steps are taken straightforward, which corresponds to the warm up stage in algorithm 1. Then the robot pushes objects using the computation result from RMPPI. From the pusher trajectories of the 'middle' target pose, we can notice that tiny direction adjustments are adopted to keep the object facing forward. Comparing the trajectory distribution towards the same target pose between (c) and (d), we can notice that the network is able to adapt itself to the real dynamics of different objects. The final poses reached are all close to the goal poses, proving that RMPPI is of high robustness in the real environment. The length of each pushing is set as 0.5 cm. Because only the first action is executed, the motion is not smooth with momentary stop after each pushing action. Executing the first two or three actions of the sequence and running the model in parallel can make the motion smooth, this will be done in the future work. Our code, model and videos are available at https://github.com/HitLyn/RMPPI.

### C. Analysis

We choose objects with different physical parameters as test objects and measure their parameters after the experiments. The results are shown in table II. Five different objects (A-E) are used to evaluate our model's adaptability, among which object A has almost the same parameters (size, mass, friction) as the prototypical object (P) we used for training the model-free baseline. Table III shows the success rate of pushing objects to target poses under different thresholds. Our model fits all five object motion properties and achieves a reasonable pushing success rate, while the model-free baseline does not work well on any object that differs from the prototypical one trained in simulation.

In order to get a comprehensive evaluation for each pushing result, we set a score calculation formula with the final error in $X$ direction, $Y$ direction, and $\theta$:

$$S = \frac{1}{E_x + 3 \cdot E_y + 0.5 \cdot E_\theta + \sigma} \quad (9)$$

in which $E_x$, $E_y$ and $E_\theta$ are the final errors to the target pose, constant $\sigma = 1e-7$. The factors used for the different errors are chosen to keep them on the same magnitude. During the real experiments, we find that the two hyperparameters in RMPPI that influence the pushing performance the most are $K$ in (3) and $T$ in algorithm 1; that is, the number of previous states used as inputs to the LSTM, and the number of timesteps used in the rollouts before calculating the costs. During the LSTM warm up stage, we simply set straight forward pushes as the first $K+1$ actions of the pusher. To find the best combination $(K^*, T^*)$, we try different groups of $K$ and $T$ in real experiments and plot part of the combined performance in Figure 7. As $K$ increases from 2 to 4, the score keeps increasing and reaches the top at 4. This means the LSTM module needs a sequence of at least 4 time steps to get the best prediction; a longer input sequence will not improve the prediction accuracy further. Because the prediction error can be accumulated through action sequence $[u_{step}, ..., u_{step+T-1}]$ from the current step to $T$ later steps, there is also the best prediction step number $T^*$. Through testing, we find the best combination is $K^* = 4, T^* = 5$.

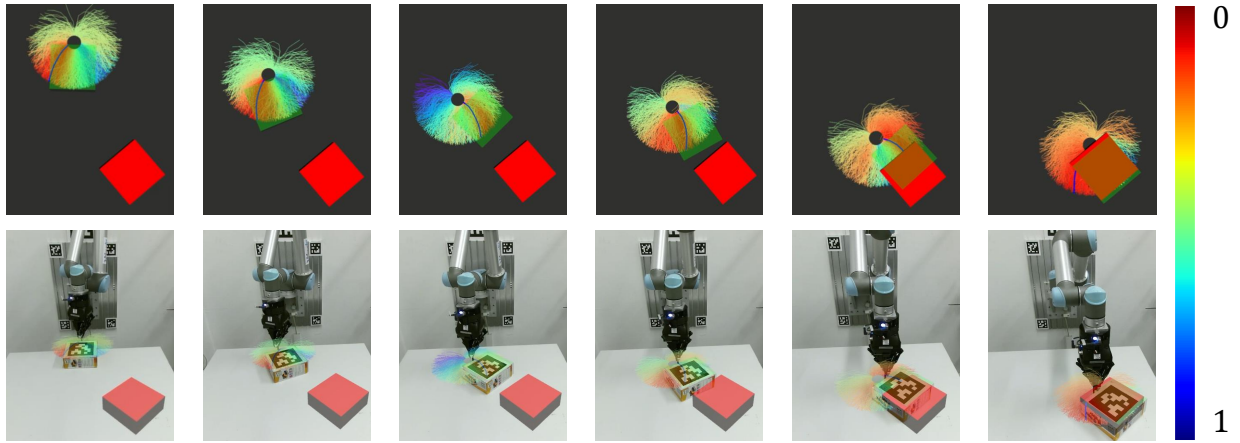

Fig. 5. Example pushing trajectory of object C. The pictures in the first row show visual markers in rviz in which the green marker represents the current pose of the object and the red marker represents the goal pose. For each pushing step, 1024 sampled rollout action sequences (pusher trajectories) are sampled. The different colors of the trajectories are normalized costs. One optimal action sequence is computed and shown by the blue line in every picture. Second row: Camera images from the real robot experiment with visualization overlays.

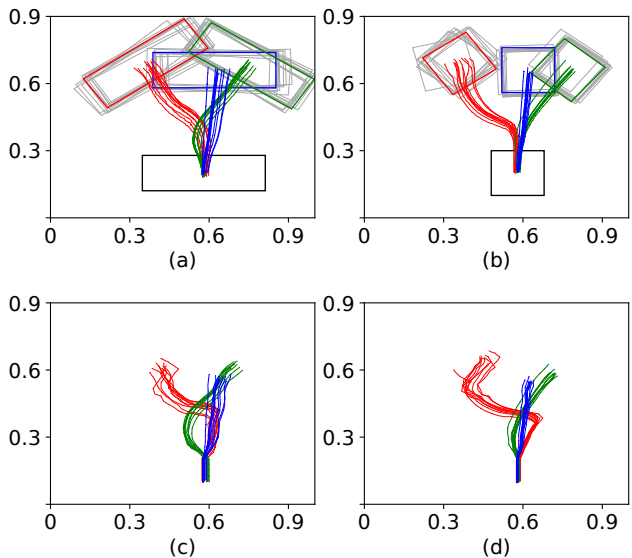

Fig. 6. Example trajectories from the real robot experiment. (a) and (b) denote trajectories of objects E and C, respectively, during the pushing process towards the 3 target poses 'left', 'middle' and 'right'. Target poses are marked by 'red', 'blue' and 'green' squares, respectively. Light gray squares represent the final poses in the real experiment. Starting positions are also denoted by black squares. (c) and (d) denote the corresponding trajectories of the pusher generated by RMPPI in the same experiment (a) and (b).

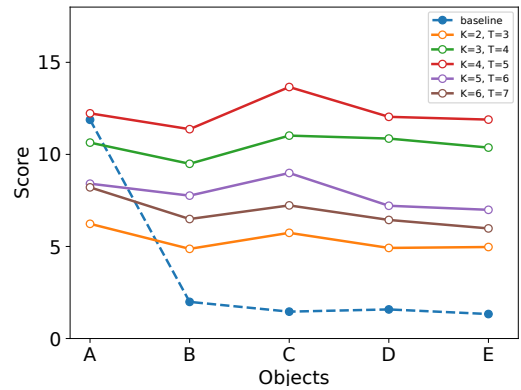

Fig. 7. The scores computed according to the evaluation formula, $K+1$ is the length of the sequence fed into LSTM for each prediction, which is also the same as that in equation (3). T is the number of steps to roll out in algorithm 1, the baseline is the score of the model-free approach.

## V. CONCLUSION

In this paper, we build a recurrent model which can adapt to the real interaction dynamics in object pushing tasks. RMPPI is proposed as the controller. Through Domain Randomization, we bridge the gap between simulation and real environment. The model is trained in simulation only but can be used in the real environment without any fine-tuning. Although the model requires an initial warm-up stage for adjusting itself, this is exactly what humans do as well when we start working with a novel object. Results show that the new algorithm is of high robustness. Through the analysis of the results, our key findings are:

- Given proper state variables, a well-trained LSTM-based model can learn to predict object motions for objects of different sizes and shapes, self-adapting to the actual object dynamics online after only a few pushing steps like a human.
- Recurrent model can be integrated effectively in an Model Predictive Control framework.
- Domain Randomization is an effective tool for bridging the gap between simulation and real environment in robotic learning tasks.

TABLE III
SUCCESS RATE COMPARISON WITH MODEL-FREE BASELINE

| Threshold | Method | A | B | C | D | E |
|---|---|---|---|---|---|---|
| $(E_x < 0.025(m)) \& (E_y < 0.010(m)) \& (E_\theta < 0.052(rad))$ | RMPPI | 85.3% | 82.8% | 87.2% | 83.3% | 81.4% |
| | Model-Free | 85.5% | 22.6% | 19.1% | 11.7% | 12.3% |
| $(E_x < 0.035(m)) \& (E_y < 0.015(m)) \& (E_\theta < 0.087(rad))$ | RMPPI | 90.5% | 88.9% | 91.5% | 88.3% | 87.5% |
| | Model-Free | 87.8% | 24.9% | 19.6% | 12.5% | 12.5% |
| $(E_x < 0.050(m)) \& (E_y < 0.025(m)) \& (E_\theta < 0.17(rad))$ | RMPPI | **93.5%** | **90.9%** | **93.9%** | **91.6%** | **89.5%** |
| | Model-Free | 92.8% | 25.6% | 19.3% | 14.8% | 13.7% |

Based on the current research, our future work will be taking images as input for the model, learning and predicting the motion property of arbitrarily shaped objects. One limitation of the proposed method is that while pushing, the robot cannot yet effectively switch sides to control the object more precisely. However, benefiting from one of the advantages of model-based control, the learned model can be adapted to new tasks. Part of our future work is to add an upper-level controller in RMPPI. By contrast, the model-free method learns to switch sides during pushing, but Automatic Domain Randomization (ADR) may be necessary during the training process to let the policy learn to push unknown objects. This will make exploration and training much more time-intensive. Another limitation is the ability of generalization to off-centered objects during real experiments, which can be learned by the model through adding more off-centered objects in simulation.

REFERENCES

[1] M. Bauza and A. Rodriguez, "A probabilistic data-driven model for planar pushing," in *2017 IEEE International Conference on Robotics and Automation (ICRA)*, 2017, pp. 3008–3015.
[2] M. Bauza, F. Alet, Y. Lin, T. Lozano-Perez, L. Kaelbling, P. Isola, and A. Rodriguez, "OmniPush: accurate, diverse, real-world dataset of pushing dynamics with RGB-D images," in *NeurIPS Physics Workshop*, 2018.
[3] K.-T. Yu, M. Bauza, N. Fazeli, and A. Rodriguez, "More than a million ways to be pushed. a high-fidelity experimental dataset of planar pushing," in *2016 IEEE/RSJ International Conference on Intelligent Robots and Systems (IROS)*, 2016, pp. 30–37.
[4] M. Andrychowicz et al., "Learning Dexterous In-Hand Manipulation," *CoRR*, 2018.
[5] I. Akkaya et al., "Solving Rubik's Cube with a Robot Hand," *arXiv:1910.07113v1*, 2019.
[6] E. F. Camacho and C. B. Alba, *Model Predictive Control*. Springer Science & Business Media, 2013.
[7] G. Williams, P. Drews, B. Goldfain, J. M. Rehg, and E. A. Theodorou, "Aggressive Driving with Model Predictive Path Integral Control," in *2016 IEEE International Conference on Robotics and Automation (ICRA)*, 2016, pp. 1433–1440.
[8] G. Williams, N. Wagener, B. Goldfain, P. Drews, J. M. Rehg, B. Boots, and E. A. Theodorou, "Information theoretic MPC for model-based reinforcement learning," in *2017 IEEE International Conference on Robotics and Automation (ICRA)*, 2017, pp. 1714–1721.
[9] M. T. Mason, "Mechanics and planning of manipulator pushing operations," *The International Journal of Robotics Research*, vol. 5, no. 3, pp. 53–71, 1986.
[10] C. E. Rasmussen, "Gaussian processes in machine learning," in *Summer School on Machine Learning*. Springer, 2003, pp. 63–71.
[11] W. Liu, Z. Wang, X. Liu, N. Zeng, Y. Liu, and F. E. Alsaadi, "A survey of deep neural network architectures and their applications," *Neurocomputing*, vol. 234, pp. 11–26, 2017.
[12] A. Rajeswaran, V. Kumar, A. Gupta, G. Vezzani, J. Schulman, E. Todorov, and S. Levine, "Learning complex dexterous manipulation with deep reinforcement learning and demonstrations," *arXiv:1709.10087*, 2017.
[13] L. Kaiser, M. Babaeizadeh, P. Milos, B. Osinski, R. H. Campbell, K. Czechowski, D. Erhan, C. Finn, P. Kozakowski, S. Levine et al., "Model-based reinforcement learning for atari," *arXiv:1903.00374*, 2019.
[14] A. Kloss, S. Schaal, and J. Bohg, "Combining learned and analytical models for predicting action effects," *arXiv:1710.04102*, 2017.
[15] M. Kopicki, S. Zurek, R. Stolkin, T. Moerwald, and J. L. Wyatt, "Learning modular and transferable forward models of the motions of push manipulated objects," *Autonomous Robots*, vol. 41, no. 5, pp. 1061–1082, 2017.
[16] M. Bauza, F. R. Hogan, and A. Rodriguez, "A data-efficient approach to precise and controlled pushing," *arXiv:1807.09904*, 2018.
[17] E. Arruda, M. J. Mathew, M. Kopicki, M. Mistry, M. Azad, and J. L. Wyatt, "Uncertainty averse pushing with model predictive path integral control," in *2017 IEEE-RAS 17th International Conference on Humanoid Robotics (Humanoids)*, 2017, pp. 497–502.
[18] J. Tobin, R. Fong, A. Ray, J. Schneider, W. Zaremba, and P. Abbeel, "Domain randomization for transferring deep neural networks from simulation to the real world," in *2017 IEEE/RSJ International Conference on Intelligent Robots and Systems (IROS)*, 2017, pp. 23–30.
[19] X. B. Peng, M. Andrychowicz, W. Zaremba, and P. Abbeel, "Sim-to-real transfer of robotic control with dynamics randomization," in *2018 IEEE International Conference on Robotics and Automation (ICRA)*, 2018, pp. 1–8.
[20] D. P. Kingma and J. Ba, "Adam: A method for stochastic optimization," *arXiv:1412.6980*, 2014.
[21] K. Lowrey, A. Rajeswaran, S. Kakade, E. Todorov, and I. Mordatch, "Plan online, learn offline: Efficient learning and exploration via model-based control," *arXiv:1811.01848*, 2018.
[22] T. P. Lillicrap, J. J. Hunt, A. Pritzel, N. Heess, T. Erez, Y. Tassa, D. Silver, and D. Wierstra, "Continuous control with deep reinforcement learning," *arXiv:1509.02971*, 2015.
[23] M. Andrychowicz, F. Wolski, A. Ray, J. Schneider, R. Fong, P. Welinder, B. McGrew, J. Tobin, O. P. Abbeel, and W. Zaremba, "Hindsight experience replay," in *Advances in Neural Information Processing Systems*, 2017, pp. 5048–5058.
[24] M. Plappert, M. Andrychowicz, A. Ray, B. McGrew, B. Baker, G. Powell, J. Schneider, J. Tobin, M. Chociej, P. Welinder et al., "Multi-goal reinforcement learning: Challenging robotics environments and request for research," *arXiv:1802.09464*, 2018.
[25] G. Brockman, V. Cheung, L. Pettersson, J. Schneider, J. Schulman, J. Tang, and W. Zaremba, "Openai gym," *arXiv:1606.01540*, 2016.
[26] E. Todorov, T. Erez, and Y. Tassa, "Mujoco: A physics engine for model-based control," in *2012 IEEE/RSJ International Conference on Intelligent Robots and Systems*, 2012, pp. 5026–5033.
[27] J. Wang and E. Olson, "AprilTag 2: Efficient and robust fiducial detection," in *2016 IEEE/RSJ International Conference on Intelligent Robots and Systems (IROS)*, 2016, pp. 4193–4198.
[28] P. Ruppel, N. Hendrich, S. Starke, and J. Zhang, "Cost functions to specify full-body motion and multi-goal manipulation tasks," in *2018 IEEE International Conference on Robotics and Automation (ICRA), Brisbane, QLD*, 2018, pp. 3152–3159.